\documentclass[12pt,a4paper]{article}
\usepackage{graphicx} 

\usepackage{amsmath} 
\usepackage[hyperindex=true, colorlinks=true, linkcolor=black, urlcolor=cyan, citecolor=cyan]{hyperref}

\title{Predicting  Air Temperature from Volumetric Urban Morphology with Machine Learning}
\author{
Berk K{\i}v{\i}lc{\i}m
and 
Patrick Erik Bradley
}
\date{}

\begin{document}

\maketitle

\vspace*{-.5cm}



%

\paragraph{Acknowledgements.}
Our sincere gratitude goes to Martin Breunig for
offering the opportunity to be part of this project, and to Markus Wilhelm
Jahn for his excellent guidance of the first author during the initial stage of this work.

\paragraph{Declaration of conflicting interest.}

\paragraph{Funding statement.}
This work
is supported by the Deutsche Forschungsgemeinschaft under project number
469999674.

\paragraph{Ethical approval and informed consent statements.}

\paragraph{Data availabilty statement.}
Data produced in this work can be provided upon request. The code for this work can be 
fund under the following link:

\url{https://github.com/BerkKivilcim/Urban-Heat-Modelling/tree/main}

\paragraph{Any other identifying information.}

\newpage

\begin{center}
{\LARGE\bf Predicting Air Temperature from Volumetric Urban Morphology with Machine Learning}
\\[5mm]
{\normalsize\today}
\end{center}

\begin{abstract}
Environmental parameters such as air temperature are critical determinants of human quality of life and energy efficiency management. Urban areas are densely populated and also highly correlated with some of these natural phenomena through urban morphology and landscape spatial patterns. Consequently, predicting the effects of urban plans on environmental parameters is essential for proper decision making and planning to enhance the living conditions of cities. Previous studies have highlighted the strong correlation between urban morphology and air temperature, underscoring the importance of employing three-dimensional data in those analyses. In this study, we firstly introduce a method that converts CityGML data into voxels which works efficiently and fast in high resolution for large scale datasets such as cities but by sacrificing some building details to overcome the limitations of previous voxelization methodologies that have been computationally intensive and inefficient at transforming large-scale urban areas into voxel representations for high resolution. Those voxelized 3D city data from multiple cities and corresponding air temperature data are used to develop a machine learning model. Before the model training, Gaussian blurring is implemented on input data to consider spatial relationships, as a result the correlation rate between air temperature and volumetric building morphology is also increased after the Gaussian blurring. After the model training, the prediction results are not just evaluated with Mean Square Error (MSE) but some image similarity metrics such as Structural Similarity Index Measure (SSIM) and Learned Perceptual Image Patch Similarity (LPIPS) that are able to detect and consider spatial relations during the evaluation process. This trained model is capable of predicting the spatial distribution of air temperature by using building volume information of corresponding pixel as input. By doing so, this research aims to assist urban planners in incorporating environmental parameters into their planning strategies, thereby facilitating more sustainable and inhabitable urban environments. 
\end{abstract}

\clearpage

\section{Introduction}
Buildings are the most active areas of human activity and have a significant impact on the urban thermal environments by altering the heat exchange \cite{han2015toward, lan2017urban, liu2020climate, ren2019capturing, zhong2017revealing}. Moreover, a significant majority of the world's population is predicted to live in urban environments in the future \cite{tehrani2024predicting}. Besides, temperature increases contribute to health issues with potential for heat-related illnesses and disrupt ecosystems and adversely affect biodiversity \cite{arnfield2003two}. Therefore, it is essential to research the urban thermal environment containing buildings \cite{yang2020application}. Although it is recognised that climate issues still have limited impact on urban planning processes \cite{eliasson2000use, ng2012towards}, it is partly because of a gap between urban planners and climatologists \cite{eliasson2000use}. To help urban planners and decision makers better understand and use the research findings, linking the climate issues to planning parameters can be more helpful than geographic or morphological parameters \cite{lan2017urban}. Therefore, in this article, a machine learning model is trained to predict urban near-surface air temperature by using building volumes. 
\\\\
The power of machine learning algorithms allows to evaluate environmental indicators on a large scale and to map urban air temperature \cite{fathi2020machine, liu2017machine, tekouabou2022reviewing, venter2020hyperlocal, yoo2018challenges}, since those algorithms have the advantage of solving complex non-linear problems with fewer computing sources and less time \cite{fan2024exploring}. Thanks to machine learning methods, we can now better model the patterns of these urban forms to refine those of future cities to meet the needs of rapid urbanization \cite{tekouabou2022reviewing}.
Although the results obtained when using the machine learning models might differ from the actual measurements while exhibiting a similar trend as the measurements, this  makes it still reasonable and acceptable \cite{lau2024investigating}. This issue  also occurred in the trained models of the present work. In the end, with the aid of trained models in this study, urban planners can manipulate the building volume values as they desire for the input of the machine learning model in order to observe how their plans will impact environmental factors. It is expected that  changing the spatial arrangement of urban components may affect the land surface energy distribution \cite{zheng2019higher}. This could allow them to balance and control their plans based on these impacts, potentially reducing the occurrence of flawed urban planning.
\\\\
Many of the previous studies focused on investigating the close relation between temperature and buildings \cite{yang2020application, hu2020modeling, li2021quantifying, oke1995heat, stewart2012local, voogt2003thermal, wu2010urban, zhou2011does}, while many others implemented machine learning models to predict LST or air temperature by considering urban morphology \cite{tehrani2024predicting, fan2024exploring, lau2024investigating, lin2024does, liu2023urban, raaymakers2024understanding}. According to these studies, a higher building volume contributes to warmer environment within the city centre. Therefore, optimisation of the building volume should be seriously considered, especially during the urban planning decision-making \cite{isa2020building}. However, most of these studies utilise the average building height of regions while focusing on selected regional blocks or local climate zones for predictions. On the other hand, our methodology incorporates building heights and footprints directly without averaging, and also associates volumes with two-dimensional air temperature raster data, enabling predictions not just for specific selected region blocks, but on a per-pixel basis. This approach will allow for very high-resolution predictions to be made swiftly as higher-resolution meteorological data become available in the future. 
\\\\
In addition, by performing pixel-level predictions, it is possible to express the results in raster format and test the model's accuracy by comparing them with ground truth data in the same format. This approach allows us to not just evaluate the models with shallow metrics like Mean Square Error (MSE), but also enables to capture spatial patterns and perform quantitative assessments similar to human visual perception using metrics such as SSIM (Structural Similarity Index) \cite{wang2004image}, and LPIPS (Learned Perceptual Image Patch Similarity) \cite{zhang2018unreasonable}. Moreover, our model's process involved data from ten different cities, with seven used in the training phase and the remaining three for testing. For this reason, the present model still provides a generalised and robust prediction capability.
\\\\
Besides, land surface temperature (LST) draws significant attention, as it modulates the air temperature of the lower layer of the urban atmosphere \cite{voogt1998effects}. However, since air temperature data over 2 meters above the surface is already provided by the German Weather Service as an open-source data, and LST is also sensitive to surface emissivity and reflectivity which might fool the trained model using many different cities, the air temperature data is used instead of LST data for the present methodology.  
\\\\
One of the greatest challenges of urban planning today is to produce urban forms that meet the challenge of today’s cities \cite{tekouabou2022reviewing}.  It is mentioned in the past studies that the landscape pattern of two-dimensional space alone is inadequate in explaining the complex thermal phenomena occurring in urban areas \cite{zheng2019higher} and the correlation between thermal phenomena and the 3D-Volume-Index was higher than the 2D-Area-Index \cite{yang2020application}. However, the role of urban morphology, such as building height is often overlooked in many cases \cite{liu2023urban}. Since height-related indicators have been typically chosen as the major parameters to characterise the three-dimensional landscape morphology \cite{cai2018water, guo2016characterizing}, one of the major limitations has been the difficulty of obtaining high-resolution 3D information about the scale of entire metropolitan areas \cite{zheng2019higher} and accurately estimating the height of buildings on a large scale to obtain the 3D structure of buildings. This is a tough challenge \cite{wu2013development}. 
In addition to that, many analyses from previous studies rely on official urban datasets provided by governments or profit-making organizations, which include building information \cite{lau2024investigating}. Therefore, available and accessible 3D urban morphology data have become essential for extensive academic research on the built environment and urban climate, and a rapid methodology for extracting urban morphology information is urgently needed \cite{ren2020developing}.
\\\\
To process the machine learning training, the voxel data type is chosen, since voxel resolution can be adaptable to any resolution. A voxel has been defined as a single primitive data element that represents the properties of real objects \cite{zlatanova2016towards}. They have definite and tangible volumes which can be used for volumetric calculations and geometric comparisons, and also offer a discrete approximation of the geometry with the advantage that individual voxels can be associated with thematic information. Each voxel can be mapped onto real-world coordinates as an unambiguous, definite representation \cite{heeramaglore2022semantically}. The usefulness for volumetric calculations is an advantage that is absent from other model types \cite{heeramaglore2022semantically}. These advantages make voxels  a perfect format for volumetric calculations of real-world buildings and make it possible to associate voxel regions with thematic temperature data for machine learning training. However, conventional methodologies for obtaining the voxel data makes the need for a large amount of laser scanning, and other sensor data are necessary \cite{pusacker2024concept}. For this reason, a CityGML-to-voxel conversion method is used, since open source CityGML data for the Thuringia/Germany region are available. CityGML is a widely used open data model based on the extensible markup language (XML), capable of describing model elements in five levels of detail \cite{padsala2021application}. 
Some of the CityGML-to-voxel conversion algorithms introduced in previous studies \cite{mulder2015automatic, nourian2016voxelization, willenborg2016semantic} 
implement only simulations on the scale of individual buildings 
instead of a whole city scale \cite{heeramaglore2022semantically, konde2017web, ridzuan20233d}. 
Cf.\ also \cite{GeoInfo2021,mdpi2022,DissMarkus}
for extraction methods of watertight volumetric models from wireframe data like CityGML, and their robustness.
However, those methodologies are based on geometric intersection procedures. This means that those algorithms would be insufficient for a direct voxelization of large 3D city datasets made up for complex geometry, so some issues need clarification about the conversion process, especially for large datasets. These conversion can only be achieved through time consuming and computationally intensive workflows \cite{heeramaglore2022semantically}.  Due to these demands on large scale voxel city models for air temperature and urban morphology relation analyses, first a methodology is introduced in this study which works fast for creating high resolution voxel data which compromises the detail level of buildings.  
\\\\
Some of the key novelties of this study includes: 1) Rapid generation of 3D raster data (voxels) from open-source CityGML data, involving the combined building footprints and volumes. 2) The integration of this 3D volumetric raster data with 2D air temperature data, facilitating the use of machine learning techniques for predicting urban heat islands while also considering the spatial adjacency of the data. 
\newline

Consequently, voxelised data can be utilised for forecasting various other natural phenomena in the future. The findings of this study are intended to provide a foundational framework for future research, in particular the ongoing research project \emph{Distributed Simulation of Processes in Buildings and City Models}, funded by the German Research Foundation (DFG), where they can provide a basis for testing mathematical simulation models.
They also offer urban planners not only insights into predicting air temperature, but also the tools to evaluate a broader spectrum of environmental parameters.

\section{Methodology}
This study exclusively utilized open-source data and open-source software tools. The employed datasets encompass CityGML data pertaining to the Thuringia state in Germany \cite{CityGML}, coupled with hourly air temperature measurements provided by the German Weather Service. The temperature data\-sets present air temperature at a height of 2 meters above ground level and feature a 1 km spatial resolution \cite{krahenmann2018high}. The air temperature data can be accessed in \cite{air_temp}. Data processing procedures were mainly conducted using Python. Additionally, Paraview \cite{Paraview} was used for some visuzalitazion tasks, while QGIS \cite{QGIS_software} and several of its plugins such as CityJSON plugin \cite{vitalis2020cityjson} and GDAL rasterize tool \cite{GDAL_rasterize}, is needed for data preparation steps to implement the voxelisation process. In addition, another open-source tool named  'citygml-tools' \cite{ledoux2019cityjson} was used for converting CityGML data into the JSON format for loading CityGML data into QGIS. 

\subsection{Study Area}
In this study, ten cities from the state of Thuringia in Germany, varying in size and population density, were selected for the analysis. Seven were used for training and three for testing. The dimensions of the selected areas for these cities are as follows: Erfurt (10km x 8km), Jena (8km x 10km), Weimar (10km x 8km), Suhl (8km x 10km), Altenburg (10km x 8km), Sondershausen (12km x 6km), Gotha (10km x 8km), Sonneberg (8km x 10km), Schmalkalden (10km x 8km), and Gera (8km x 10km). When associating CityGML data with air temperature measurements, the air temperature datasets used were those recorded in the same year that the CityGML datasets were created for each respective city. This approach was implemented to minimize inconsistencies arising from temporal resolution discrepancies. To present the results more clearly and to better highlight the advantages of our methodology, instead of using multi-temporal datasets, we utilised air temperature data by averaging the values obtained specifically at 01:00 AM during the month of July. The rationale for selecting this particular time is based on previous studies, which demonstrated that the correlation between air temperature and urban morphology reaches its maximum at 01:00 AM during the summer \cite{lan2017urban}. This high correlation has made it possible to present and interpret the results in a clearer and more comprehensible manner.
\\\\
The coordinate system of the CityGML data, which was used in this research, is EPSG:25832, while the air temperature data possesses latitude and longitude coordinates under the EPSG:4326 system, in addition to X and Y coordinates under the EPSG:3034 system. 
\\\\
The air temperature data, covering all of Germany, was cropped using the upper right and lower left coordinates of the CityGML data to align two datasets accurately. The consistency of data alignment following the crop ping process was assessed by converting the air temperature data into raster format from netCDF format and subsequently loading it into QGIS. The alignment was checked through qualitative comparisons between CityGML region and cropped air temperature region within a shared coordinate system in QGIS to confirm consistency. The visual representation of this comparisons given in Figure S1. 

\subsection{CityGML to Voxel Conversion}
\begin{figure}[h!]
\centering
\includegraphics[width=1\textwidth]{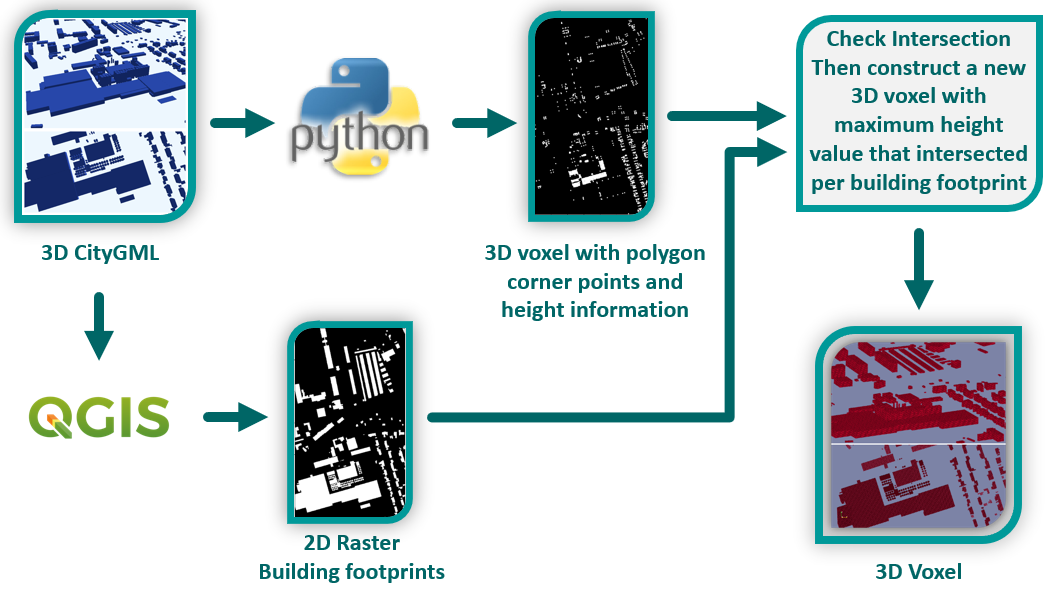}
\caption{Workflow of the voxelisation process}
\label{fig:voxelization}
\end{figure}

Traditional techniques for converting CityGML data into voxels operate by calculating intersections between CityGML and potentially billions of grid points for high-resolution and extensive areas. Although this method can model the many details of buildings and produce complex building voxels, it requires substantial computational power and time. Our method simplifies the process significantly by focusing solely on regions with buildings within a two-dimensional plane, assigning a single height value to each building, thereby enabling the rapid construction of less detailed buildings. Consequently, the first step in our approach is to derive a 2D raster dataset from CityGML data. A brief workflow illustration is given in Figure \ref{fig:voxelization}.
\vspace*{0.1cm}

\subsubsection{Retrieving 2D Building Footprint Areas}
Initially, the CityGML data, downloaded via \cite{CityGML}, covered an area of 2 km $\times$ 2 km. Therefore, these data were merged to create a single comprehensive CityGML file for each city. This resulting CityGML file was then converted into the CityJSON format and imported into QGIS using the “CityJSON Loader” plugin. Subsequently, GDAL’s rasterisation tool was employed to produce raster data for any desired region at any specified resolution. In this study, the raster resolution was set at 1 metre, and the necessary boundary regions for the raster image were extracted from the CityGML data. Additional parameters selected during the use of the tool included: \textit{“A fixed value to burn: 1”, “Assign a specified no data value to output bands: -999”, “Output data type: Int16”, “Pre-initialize the output image with value: 0”.} The output of this process is extracted as GeoTiff format in the coordinate system of EPSG:25832. The illustration of those extracted 2D raster dataset is given in Figure S2.

\subsubsection{Real-world coordinate system to voxel coordinate system}
This step involves calculating which location indices in our voxel system correspond to each building's polygon vertices that possess EPSG:25832 coordinate data. In addition to the horizontal plane coordinates of these corner points, the data also includes building height information. Consequently, this allows us to utilise the heights of buildings to combine them with 2D building footprints from the raster image. 
\\\\
For instance, when the region encompassing the city of Erfurt with an area of 10 km $\times$ 8km, and the voxel resolution is selected as 1 metre, the number of  voxels in the horizontal plane should be 10000 x 8000. Consequently, the indices of the voxel array range from 0 to 9999 for width and from 0 to 7999 for height in Python indexing. Considering all these factors, a normalisation method was employed to transform data from the EPSG:25832 coordinate system to the local voxel coordinate system. The formulas used for the X and Y axes are provided below, cf.\ eq.\ (\ref{voxelCoordX}) and (\ref{voxelCoordY}). The reason for employing different formulas for the X and Y axes is the orientation of arrays in Python, where the origin (0,0) index is at the top-left corner, whereas the real-world coordinate system of the study area places the origin at the bottom-left. This discrepancy causes a flip along the Y-axis, leading to inconsistencies. The formula used for the Y-axis adjusts this issue. Furthermore, after the normalisation process in the formula, the resulting values between 0 and 1 are multiplied by the width or height values using $width - 1$ or $height - 1$. This minus 1 subtraction adjustment is made because Python indexing starts at 0. In the end, as voxels constitute discrete grids, the new coordinates derived from the formula must be integers. Therefore, a rounding operation is applied to the computed values to ensure they conform to this requirement.

\begin{align}\label{voxelCoordX}
X_{\text{voxel}}& = \displaystyle \text{round} \bigg( \frac{(X_{\text{i}} - X_{\text{min}})}{(X_{\text{max}} - X_{\text{min}})} \cdot (width - 1) \bigg) 
\\\label{voxelCoordY}
Y_{\text{voxel}} &= \displaystyle \text{round} \bigg( \frac{(Y_{\text{max}} - Y_{\text{i}})}{(Y_{\text{max}} - Y_{\text{min}})} \cdot (height - 1) \bigg)
\end{align}

\subsubsection{Assigning the height information to building footprints}

\begin{figure}[h]
\centering
\includegraphics[width=1\textwidth]{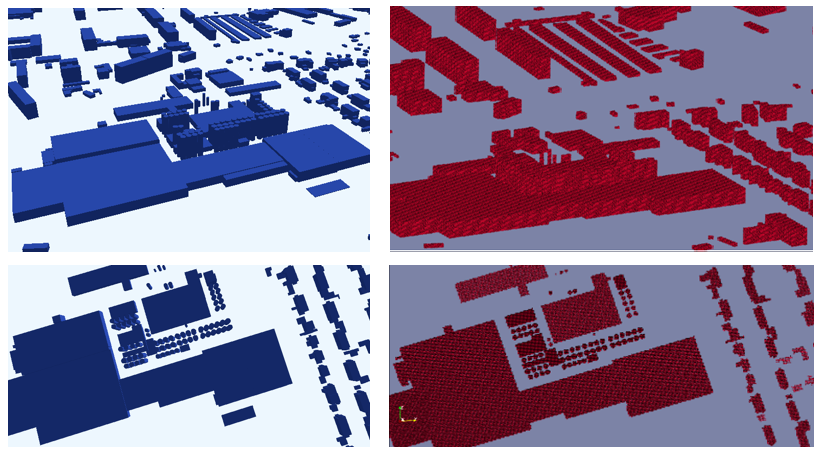}
\caption{The left column represents the CityGML visualization of Gotha from different viewing angles, the right column represents the voxel visualisation of the same region of Gotha}
\label{fig:voxel_result}
\end{figure}

In the final stage of the voxelisation process, a new voxel is generated based on the normalised building polygon vertex coordinates positioned within the voxel grid and 2D building footprint raster image. If these corner points on the voxels align with a building depicted in a two-dimensional raster image, then the height of the building is derived from the highest height value among the matching vertex coordinates. This method produces buildings in the voxel space that are highly detailed and accurate in terms of their footprint, yet adopt a simplified approach for height representation by assigning a single height value per building. Importantly, even when high-resolution voxels are desired, the process outlined here is implemented through a single for-loop that iterates only once per building which is independent from the total number of pixels in the image. This method minimises the impact on processing speed, despite the potential for a large number of pixels and high resolution. A result of this voxelisation process is demonstrated in the Figure \ref{fig:voxel_result}.

\subsection{Machine Learning Training}

\begin{figure}[h]
\centering
\includegraphics[width=1\textwidth]{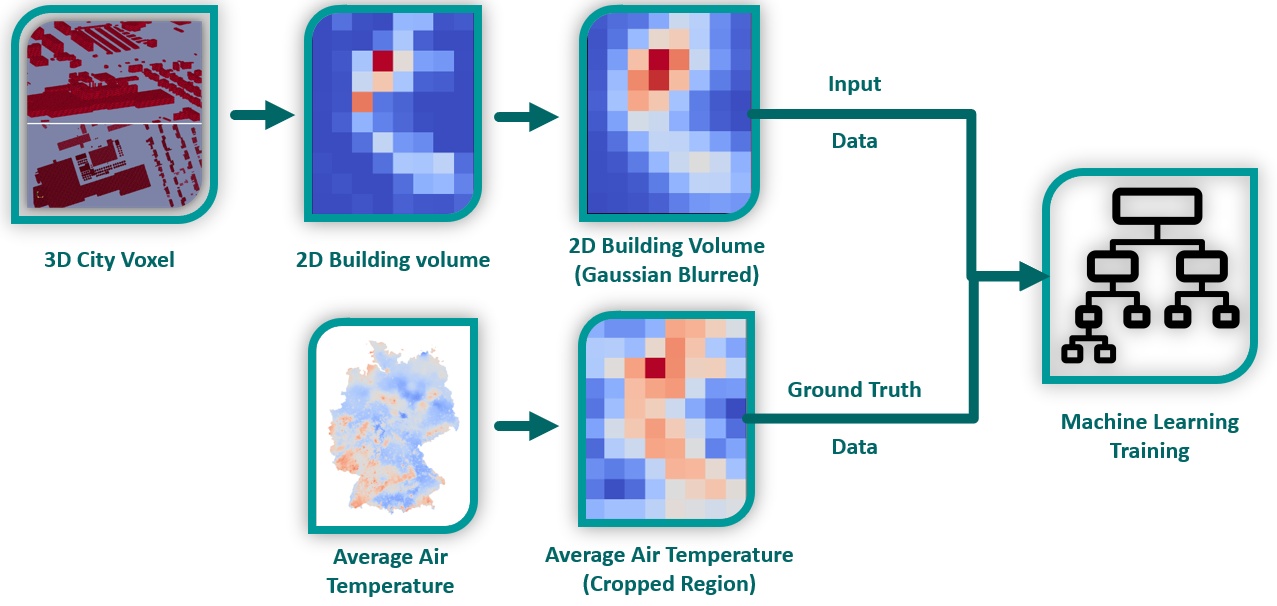}
\caption{A brief workflow demonstration of data pre-processing steps for the machine learning training}
\label{fig:ML_workflow}
\end{figure}

In the machine learning model, the phenomenon targeted for prediction is air temperature. Therefore, air temperature data with a resolution of 1km $\times$ 1km has been employed as ground truth data in the training phase. Given that the resolution of these air temperatures is 1km $\times$ 1km, the area covered by the voxels is divided into a grid commensurable with the voxel numbers. For example, the region with size of 10km $\times$ 8km is divided into a 10 $\times$ 8 grid. Afterwards,  new two-dimensional building volume data, containing the total building volume for each grid are generated. Consequently, the adjustable resolution of the voxels allows  to associate them with higher resolution air temperature data, if available. Thereby our methodology also ensures that the voxelised methods can adapt to improved or different meteorological data resolutions. After the voxelisation steps, the other implemented steps before the machine learning training is presented in Figure \ref{fig:ML_workflow}.

\begin{figure}[h!]
\centering
\includegraphics[width=1\textwidth]{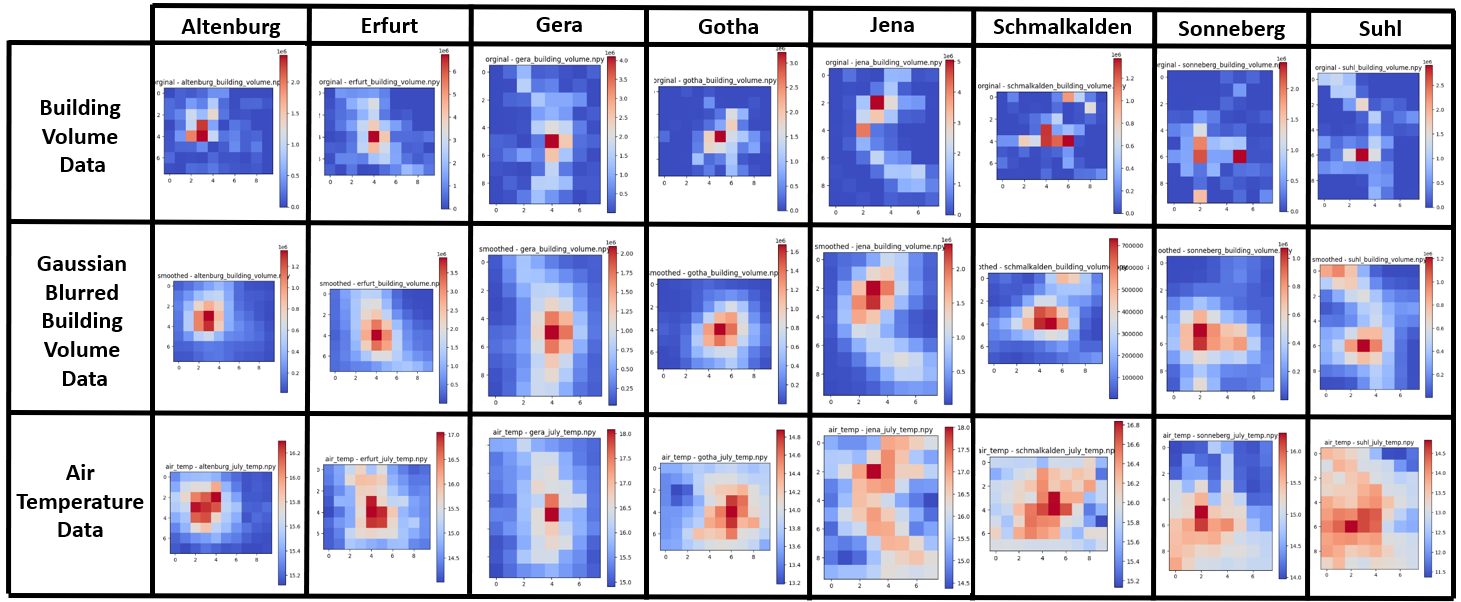}
\caption{This figure illustrates the spatial distribution of building volumes (expressed in units of $\mathrm{m}^3$) and a monthly average air temperature of July at 01:00 AM (expressed in units of $\mathrm{C}^\circ$) in selected urban areas to show the effect of Gaussian blur, and the relation between urban morphology and air temperature.}
\label{fig:data_visualization}
\end{figure}

Incorporating spatial neighbourhood characteristics during the training phase is crucial, especially when considering urban environments with varying spatial patterns since even if a region has a high building volume but is isolated without many surrounding buildings, it may exhibit a lower urban heat island effect than expected. To account for this, a Gaussian blurring method has been applied to the 2D building volume input data. This approach has demonstrated an increase in the correlation between the building volume in each city and their air temperature. The selected Gaussian kernel parameters are; \textit{sigma value = 0.85} and \textit{radius = 1}. Furthermore, this correlation tends to rise in cities with higher population densities, indicating a significant interplay between urban morphology and thermal behavior. The amount of correlation of original data and Gaussian blurred data with air temperature is given in Table \ref{tab:correlation}, while the visualisations of those datasets are presented in Figure \ref{fig:data_visualization}, which clearly shows the effect of the Gaussian blur.
\\\\
The Random Forest (RF) and Extreme Gradient Boosting (XGBoost) techniques are selected with hyper-parameter optimisation conducted via a trial and error method for machine learning training. A previous study  demonstrated the effectiveness of XGBoost compared to other techniques for predicting urban heat island effects \cite{tanoori2024machine}. For the RF model, the hyper-parameters were established as follows: \textit{number of trees} = 100,000, \textit{maximum depth of trees} = 3, \textit{minimum number of samples required to split a node} = 4, \textit{minimum number of samples per leaf} = 2, and \textit{max\_features} set to ‘sqrt’. For the XGBoost model, the parameters were set to: \textit{number of trees} = 300,000, \textit{maximum depth of trees} = 3, and \textit{learning rate} = 0.000003. Additionally, to mitigate overfitting, augmented data were utilized during training, with parameters specified as \textit{number of samples} = 100 and \textit{noise level} = 0.01."

\begin{table}[h]
\centering
\caption{The table represents the correlation values of the original 2D building volume data and the Gaussian blurred data with the air temperature}
\label{tab:correlation}
\vspace{2mm}
\begin{tabular}{|l|c|c|c|} 
\hline
Datasets & \begin{tabular}[c]{@{}c@{}}Correlation with\\ building volume\\and air temperature \end{tabular} & \begin{tabular}[c]{@{}c@{}}Correlation with Gaussian\\blurred building volume\\and air temperature\end{tabular}\\
\hline
\textbf{Altenburg}   & 0.78 & 0.93\\
\hline
\textbf{Erfurt}   &    0.85           &  0.90   \\
\hline
\textbf{Gera}         &    0.76          &  0.87   \\
\hline
\textbf{Gotha}        &     0.71         &  0.83    \\
\hline
\textbf{Jena}         &   0.73           & 0.82   \\
\hline
\textbf{Schmalkalden}         &   0.48           & 0.65  \\
\hline
\textbf{Sondershausen}         &   0.62           & 0.71 \\
\hline
\textbf{Sonneberg}         &   0.53           & 0.74 \\
\hline
\textbf{Suhl}         &   0.52           & 0.66 \\
\hline
\textbf{Weimar}         &   0.67           & 0.77   \\
\hline
\end{tabular}
\end{table} 

\clearpage
\section{Results}
\begin{figure}[h!]
\centering
\includegraphics[width=1\textwidth]{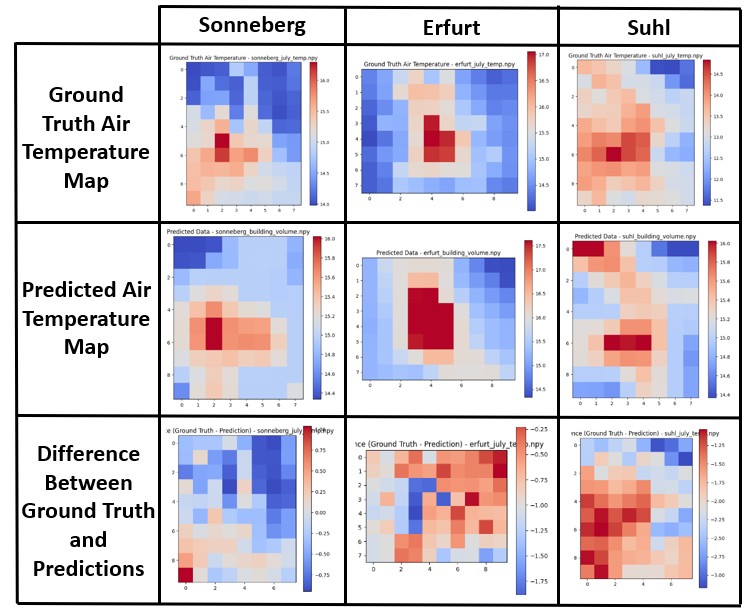}
\caption{Comparison of the air temperature prediction results obtained from the model trained with Random Forest technique and ground truth data for the test dataset which is not included during the training process. The difference map between prediction and ground truth images given in the bottom row.}
\label{fig:RF_test}
\end{figure}

\begin{figure}[h!]
\centering
\includegraphics[width=1\textwidth]{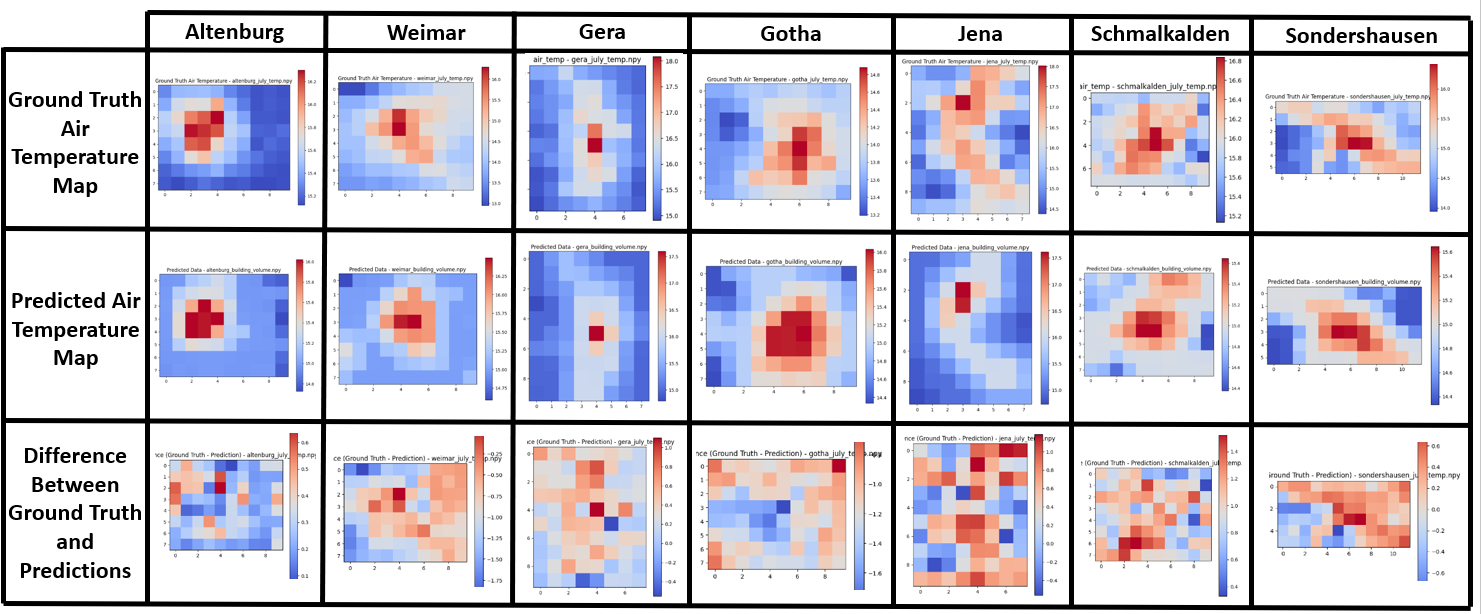}
\caption{Comparison of the predictions obtained from the model trained with Random Forest method and ground truth data for the training dataset. The difference map between prediction and ground truth images given in the bottom row.}
\label{fig:RF_train}
\end{figure}

\begin{figure}[h!]
\centering
\includegraphics[width=1\textwidth]{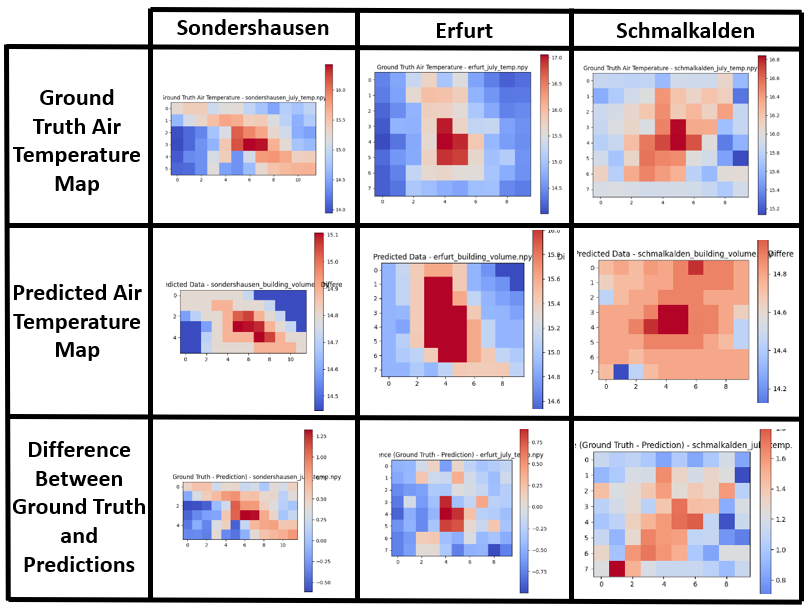}
\caption{Comparison of the air temperature prediction results obtained from the model trained with XGBoost technique and ground truth data for the test dataset which is not included during the training process. The difference map between prediction and ground truth images given in the bottom row.}
\label{fig:XGB_test}
\end{figure}

\begin{figure}[h!]
\centering
\includegraphics[width=1\textwidth]{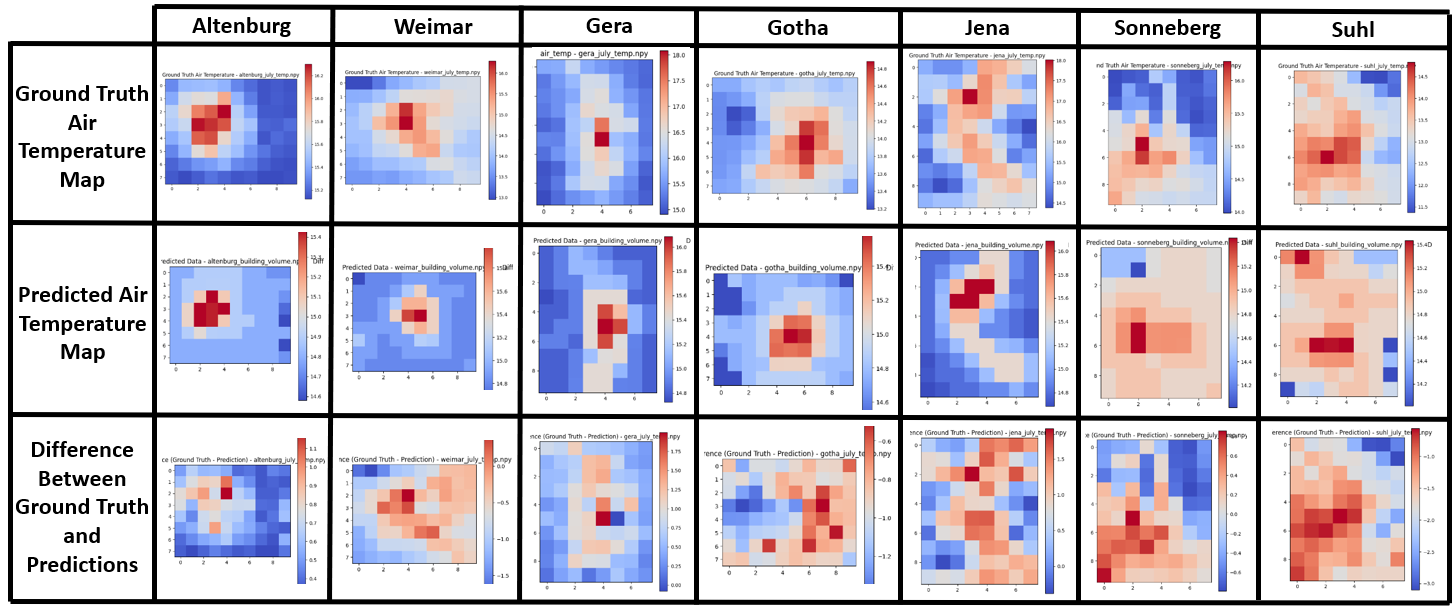}
\caption{Comparison of the predictions obtained from the model trained with XGBoost method and ground truth data for the training dataset. The difference map between prediction and ground truth images given in the bottom row.}
\label{fig:XGB_train}
\end{figure}

Since the accuracy of the trained model and the corresponding parameter selections were determined through a trial-and-error approach, multiple training processes were repeated. The resulting training outcomes were analysed both qualitatively and quantitatively.  However, in this results section, the model that qualitatively provided the most accurate results and best represented the spatial patterns of the urban heat island effect is presented. 
\\\\
For this reason, rather than focusing solely on achieving a lower MSE (Mean Square Error), visual results were used as the primary basis for accuracy. For instance, in evaluations using deeper trees with both the XGBoost and Random Forest methods, MSE values as low as 0.20 $^\circ\text{C}^2$ were achieved for XGBoost, while values around 0.45 $^\circ\text{C}^2$ were observed for Random Forest during our experiments. However, upon reviewing the qualitative results, it became evident that these seemingly satisfactory quantitative outcomes were the result of overfitting. This was especially apparent in the visual outcomes, where the urban heat island patterns of cities were not predicted in a consistent and semantically correct manner, failing to capture the expected spatial patterns. On the other hand, some experiments provide appropriate and consistent visual patterns, despite having higher quantitative error values such as 0.92 $^\circ\text{C}^2$ MSE for the Random Forest and 0.84 $^\circ\text{C}^2$ for the XGBoost techniques. 
\\\\
In addition to the visual analysis of the predictions, the differences between the predicted results and ground truth data were also examined. Upon investigation, it was observed that the Random Forest model’s error behavior did not follow any discernible spatial patterns. Instead, the error appeared to occur randomly across the spatial domain, suggesting that the model’s inaccuracies were distributed without any systematic bias or spatial structure. On the other hand, when examining the error distribution of the model trained with XGBoost, it was observed that the highest error levels were concentrated in regions corresponding to urban areas. This phenomena can be seen in the following figures of this section. 
\\\\
The comparison between the prediction results and ground truth data for the training cities (Jena, Weimar, Sondershausen, Altenburg, Gotha, Schmalkalden, Gera) obtained using the Random Forest model is presented in Figure \ref{fig:RF_train}. Similarly, the same comparison for the test cities (Erfurt, Suhl, Sonneberg) is shown in Figure \ref{fig:RF_test}. Although the spatial patterns are well predicted and presented in these Figure \ref{fig:RF_test} and Figure \ref{fig:RF_train}, some deviations in air temperature predictions were observed especially for the cities of Schmalkalden and Suhl. For instance, the air temperature predictions for Schmalkalden ranged between 14.37°C and 15.64°C, whereas the ground truth data showed a range of 15.14°C to 16.84°C. Similarly, for Suhl, the predictions fell between 14.36°C and 16.02°C, while the actual measurements ranged from 11.37°C to 14.84°C. These discrepancies directly contribute to higher MSE values. However, for other cities, significant differences in air temperature predictions were not observed. For example, the prediction range for Altenburg was 14.72°C to 16.02°C, compared to the ground truth range of 15.12°C to 16.3°C. For Erfurt, the prediction range was 14.33°C to 17.61°C, while the ground truth range was 14.01°C to 17.05°C. Similarly, Gera's prediction interval was 14.79°C to 17.59°C, with a ground truth interval of 14.90°C to 18.08°C; Gotha's prediction interval was 14.33°C to 16.03°C, compared to the ground truth of 13.19°C to 14.88°C; Jena's prediction interval was 14.72°C to 17.61°C, with the ground truth range at 14.37°C to 18.01°C; Sondershausen's prediction interval was 14.33°C to 15.64°C, while the ground truth ranged from 13.94°C to 16.42°C; Sonneberg's prediction interval was 14.33°C to 16.02°C, compared to the ground truth interval of 13.98°C to 16.29°C; and finally, Weimar's prediction interval was 14.58°C to 16.49°C, while the ground truth ranged from 12.94°C to 16.29°C.
\\\\
The similar comparison between the predictions obtained using the XGBoost model and ground truth data for the training cities (Jena, Sonneberg, Weimar, Altenburg, Gotha, Gera, Suhl) is presented in Figure \ref{fig:XGB_train}, while the comparison for the test cities (Sondershausen, Schmalkalden, Erfurt) is given in Figure \ref{fig:XGB_test}. The air temperature prediction ranges for each city using the model trained with the XGBoost method are as follows: Altenburg: 14.58°C to 15.43°C, Erfurt: 14.44°C to 16.08°C, Gera: 14.72°C to 16.08°C, Gotha: 14.44°C to 15.60°C, Jena: 14.68°C to 16.08°C, Schmalkalden: 14.01°C to 15.10°C, Sondershausen: 14.46°C to 15.10°C, Sonneberg: 14.01°C to 15.43°C, Suhl: 14.01°C to 15.43°C, and Weimar: 14.58°C to 16.08°C. The ground truth air temperature intervals for comparison have been provided in the previous paragraph.
\\\\
Considering the experiments conducted and the quantitative and qualitative results obtained, it has been demonstrated that metrics such as MSE do not fully reflect the accuracy of the model. It was also observed that hyper-parameter selection plays a critical and direct role in model performance. Additionally, the Random Forest technique was shown to accurately predict overall spatial patterns even on unseen data during training.
\\\\
In addition to the MSE comparisons, the similarities between the predicted and ground truth images were analyzed using SSIM (Structural Similarity Index), and LPIPS (Learned Perceptual Image Patch Similarity) metrics. The SSIM value of 1 indicates two images are exactly the same while 0 means no similarity. LPIPS is specifically designed to evaluate the similarities like human visual perception through deep learning techniques to overcome the shallowness of SSIM \cite{zhang2018unreasonable}. If a LPIPS value is closer to 0 that means two images are very similar to each other. The kernel size for the SSIM calculation was set to 5x5. For the LPIPS computation, all prediction and ground truth data were resized to 64x64, and the AlexNet architecture \cite{krizhevsky2012imagenet} was used as the multi-layer perceptron model within the deep neural network. That means both of these methods uses kernels to detect the spatial relationships during the evaluations. The results of these metrics are given in the Table \ref{tab:metrics}. While the model trained using the XGBoost method outperformed the Random Forest model in terms of the MSE metric, both qualitative assessments and other quantitative metrics like SSIM and LPIPS indicated that the Random Forest-based method outperformed the XGBoost model across all cities. 

\begin{table}[h]
\centering
\caption{The table represents SSIM and LPIPS values for Random Forest and XGBoost models across different cities}
\label{tab:metrics}
\vspace{2mm}
\begin{tabular}{|l|c|c|c|c|} 
\hline
\textbf{Dataset} & \multicolumn{2}{c|}{\textbf{Random Forest}} & \multicolumn{2}{c|}{\textbf{XGBoost}} \\ \cline{2-5} 
                               & \textbf{SSIM} & \textbf{LPIPS} & \textbf{SSIM} & \textbf{LPIPS} \\
\hline
\textbf{Altenburg}             & 0.89          & 0.0000059      & 0.74         & 0.0000183         \\
\hline
\textbf{Erfurt}                & 0.82          & 0.0000504      & 0.71         & 0.0002            \\
\hline
\textbf{Gera}                  & 0.88          & 0.0000377      & 0.72         & 0.0001            \\
\hline
\textbf{Gotha}                 & 0.80          & 0.0000479      & 0.77         & 0.0000430         \\
\hline
\textbf{Jena}                  & 0.77          & 0.00014        & 0.61         & 0.0003            \\
\hline
\textbf{Schmalkalden}          & 0.70          & 0.0000479      & 0.47         & 0.0001            \\
\hline
\textbf{Sondershausen}         & 0.65          & 0.0001         & 0.45         & 0.0002            \\
\hline
\textbf{Sonneberg}             & 0.65          & 0.0001         & 0.46         & 0.0001            \\
\hline
\textbf{Suhl}                  & 0.62          & 0.0001         & 0.46         & 0.0002            \\
\hline
\textbf{Weimar}                & 0.80          & 0.0000924      & 0.71         & 0.0001            \\
\hline
\end{tabular}
\end{table}

\clearpage
\section{Discussion}
Although the voxelization methodology presented in this study is designed to yield quick and efficient results, it is previously mentioned that the results compromise the level of detail of buildings. Since only a single height value is assigned per building, this approach can present challenges with more complex building structures. For instance, as clearly shown in Figure S3, the Jena Tower and the adjacent building comprise multiple components with varying height values within a single polygon component. In such cases, the voxelization process may lack sufficient precision, leading to some degree of compromise in accurately capturing the architectural complexity of certain buildings.
\\\\
While generating voxels, the intersection between 2D footprints of the buildings and the building polygon corner points in the voxel coordinate system is used. To increase the possibility of intersection, instead of treating these building polygon corner points as single pixels, 3x3 patches are used. It is important to note that the choice of the patch size is related to the resolution being used. For instance, at lower voxel resolutions, the use of patches may not be necessary, whereas at higher resolutions, larger patches might be required to ensure intersection.
On the other hand, using larger patches reduce the precision in cases where buildings are densely situated and have varying heights. However, since the voxel resolution is selected as 1 meter in this study, the use of $3\times 3$ meter patches does not significantly cause the issues in this context. Moreover, buildings that are closely positioned in many urban areas often share the similar height values which minimize the potential impact on precision. Nevertheless, even working with  1 meter resolution voxels and using $3\times 3$ patches, some buildings were not generated due to the failure to achieve intersections. For example, 24990 out of 26642 buildings were successfully matched and generated in Erfurt, it is 19154 out of 20180 in Jena, 17313 out of 18353 in Weimar, 17551 out of 18626 in Suhl, 11526 out of 13001 in Altenburg, 7803 out of 8454 in Sondershausen, 12782 out of 13620 in Gotha, 12002 out of 12814 in Sonneberg, 9847 out of 10670 in Schmalkalden, 19693 out of 22255 in Gera. 
\\\\
In addition, an alternative method is tested for generating voxels that eliminates the need for the 2D footprint image and manually defined patch sizes. Since the corner coordinates of each building is known, polygons are created by using these corner points. Then each grid in the voxels are checked to determine whether a grid is located inside a polygon or not. However, this method produced inferior results in terms of both accuracy and computational efficiency. As the total number of voxels increased, determining whether each voxel was inside a polygon or not, became increasingly time consuming. That means this alternative method lost efficiency with higher resolution voxels. Additionally, the \textit{"mpath.Path"} function used in Python to create polygons from corner coordinates failed to handle certain complex polygon structures. Figure S4 illustrates the orginal footprint image, the footprint generated using the intersection method that introduced in this paper and the building footprint produced by determining polygons based on corner points without the assistance of 2D footprint data.  
\\\\
Since the correlation between air temperature and building volume data is not perfect, it is not expected to achieve entirely accurate prediction results. However, given that cities with high population densities tend to exhibit stronger correlations with air temperature data, focusing the training process only by using cities such as Tokyo, New York, Istanbul or other cities with high population density may lead to more consistent model training. This approach could improve the performance of models designed for future analyses of metropolitan urban environments. Also, it is possible to obtain significantly different model outcomes even when using the same hyper-parameters. Therefore, it remains feasible to train models that outperform or underperform the ones presented in this study by utilizing the same hyper-parameters through trial-and-error processes across multiple trainings. 
\\\\
Besides, the methodology presented in this study not just solely focuses on the quantitative accuracy of predictions, but also observed the spatial patterns thanks to the voxels and their association with thematic data. This revealed that low MSE values alone are not sufficient to understand model accuracy. Therefore, the use of additional metrics (SSIM, LPIPS) for comparing prediction and ground truth images after post-training are useful. Among the other metrics, LPIPS stands as one of the most state-of-the-art but there are still some drawbacks of this metric. A previous study indicates that LPIPS is susceptible to such imperceptible adversarial perturbations 
where the LPIPS values are significantly affected by adding some noise or manipulating just a single pixel \cite{ghildyal2023attacking}. 

\clearpage
\section{Conclusion}
In this study, the voxel data and their representation in raster format allowed for calculating the amount building volumes for each air temperature pixel region. Since these building volumes are represented in raster format, it becomes possible to apply image processing techniques such as Gaussian blurring. Gaussian blurring enabled the integration of spatial neighborhood relationships before the training process, as the value of a pixel is influenced by adjacent pixel values. Thereby, the correlation rate is also increased between air temperature data and building volume data after implementing Gaussian blur.
\\\\
Additionally, the proposed CityGML-to-voxel conversion method facilitated the rapid generation of city-scale 3D volumetric data, accelerating experiments by allowing quick acquisition of voxel data from different regions. This rapid conversion enables urban planners to quickly implement their plans in digital applications and observe the impact on environmental indicators.
\\\\
In addition, as demonstrated by previous studies, when model accuracy is evaluated using the MSE metric, the XGBoost method produces models with low error rates. However, thanks to the raster format of our voxelized methodology, these errors are observed to be systematic and the spatial distributions are not well captured. On the other hand, the Random Forest method has higher MSE values, it qualitatively demonstrates more consistent spatial patterns and the error distribution appears more randomly.
Furthermore, the raster-based format of both predictions and ground truth data allowed for the implementations of image similarity metrics like SSIM or LPIPS that also capture and consider spatial relationships for evaluation process. Since Random Forest method qualitatively provided better results than XGBoost method, these image similarity metrics proves that qualitative finding in a quantitative manner. The SSIM and LPIPS scores for each city, indicating that the prediction results of the Random Forest method are better than XGBoost. Besides, using the testing data confirms that these accurate patterns are not due to overfitting but rather indicate good generalization.
\\\\
Taking all of these into consideration, it has been demonstrated that previous studies relying solely on MSE to evaluate their models are inadequate and lack depth. The voxel-based approach presented here takes into account spatial neighborhood relationships both before and after training, resulting in more accurate outcomes and insightful analyses. The models used in this study kept simple without any multi-variable or multi-temporal sources to highlight the benefits of the voxelized methodology, which is intended to serve as a foundation for future research and has potential applications in a variety of fields.




\bibliographystyle{plain}
\bibliography{biblio}

\begin{thebibliography}{10}

\bibitem{air_temp}
Dwd climate data center (cdc): Annual mean of station observations of daily air
  temperature at 2 meter above ground in °c for germany.
\newblock
  \url{https://opendata.dwd.de/climate_environment/CDC/grids_germany/hourly/hostrada/air_temperature_mean/}.

\bibitem{CityGML}
Open source citygml data of thuringia/germany.
\newblock
  \url{https://geoportal.thueringen.de/gdi-th/download-offene-geodaten/download-3d-gebaeudedaten}.

\bibitem{Paraview}
Sandia national labs, kitware inc, and los alamos national labs. paraview:
  Parallel visualization application.
\newblock \url{https://www.paraview.org/}.

\bibitem{arnfield2003two}
A~John Arnfield.
\newblock Two decades of urban climate research: a review of turbulence,
  exchanges of energy and water, and the urban heat island.
\newblock {\em International Journal of Climatology: a Journal of the Royal
  Meteorological Society}, 23(1):1--26, 2003.

\bibitem{cai2018water}
Zhi Cai, Guifeng Han, and Mingchun Chen.
\newblock Do water bodies play an important role in the relationship between
  urban form and land surface temperature?
\newblock {\em Sustainable cities and society}, 39:487--498, 2018.

\bibitem{eliasson2000use}
Ingeg{\"a}rd Eliasson.
\newblock The use of climate knowledge in urban planning.
\newblock {\em Landscape and urban planning}, 48(1-2):31--44, 2000.

\bibitem{fan2024exploring}
Chengliang Fan, Binwei Zou, Jianjun Li, Mo~Wang, Yundan Liao, and Xiaoqing
  Zhou.
\newblock Exploring the relationship between air temperature and urban
  morphology factors using machine learning under local climate zones.
\newblock {\em Case Studies in Thermal Engineering}, 55:104151, 2024.

\bibitem{fathi2020machine}
Soheil Fathi, Ravi Srinivasan, Andriel Fenner, and Sahand Fathi.
\newblock Machine learning applications in urban building energy performance
  forecasting: A systematic review.
\newblock {\em Renewable and Sustainable Energy Reviews}, 133:110287, 2020.

\bibitem{GDAL_rasterize}
{GDAL/OGR contributors}.
\newblock {GDAL/OGR} geospatial data abstraction software library, 2024.

\bibitem{ghildyal2023attacking}
Abhijay Ghildyal and Feng Liu.
\newblock Attacking perceptual similarity metrics.
\newblock {\em arXiv preprint arXiv:2305.08840}, 2023.

\bibitem{guo2016characterizing}
Guanhua Guo, Xiaoqing Zhou, Zhifeng Wu, Rongbo Xiao, and Yingbiao Chen.
\newblock Characterizing the impact of urban morphology heterogeneity on land
  surface temperature in guangzhou, china.
\newblock {\em Environmental Modelling \& Software}, 84:427--439, 2016.

\bibitem{han2015toward}
Yilong Han, John~E Taylor, and Anna~Laura Pisello.
\newblock Toward mitigating urban heat island effects: Investigating the
  thermal-energy impact of bio-inspired retro-reflective building envelopes in
  dense urban settings.
\newblock {\em Energy and Buildings}, 102:380--389, 2015.

\bibitem{heeramaglore2022semantically}
Medhini Heeramaglore and Thomas~H Kolbe.
\newblock Semantically enriched voxels as a common representation for
  comparison and evaluation of 3d building models.
\newblock {\em ISPRS Annals of the Photogrammetry, Remote Sensing and Spatial
  Information Sciences}, 10:89--96, 2022.

\bibitem{hu2020modeling}
Yunfeng Hu, Zhaoxin Dai, and Jean-Michel Guldmann.
\newblock Modeling the impact of 2d/3d urban indicators on the urban heat
  island over different seasons: A boosted regression tree approach.
\newblock {\em Journal of environmental management}, 266:110424, 2020.

\bibitem{isa2020building}
Nurul~Amirah Isa, Siti~Aekbal Salleh, Wan Mohd Naim~Wan Mohd, Andy Chan, Maggie
  Chel~Gee Ooi, Nur~Hidayah Zakaria, and Md~Amirul Islam.
\newblock Building volume effects on ambient temperature in the kuala lumpur
  city.
\newblock In {\em IOP Conference Series: Earth and Environmental Science},
  volume 489, page 012011. IOP Publishing, 2020.

\bibitem{DissMarkus}
M.~Jahn.
\newblock {\em Distributed \& Parallel Data Management to Support
  Geo-Scientific Simulation Implementations}.
\newblock PhD thesis, Karlsruhe Institute of Technology, 2022.

\bibitem{GeoInfo2021}
M.W. Jahn and P.E. Bradley.
\newblock Computing watertight volumetric models from boundary representations
  to ensure consistent topological operations.
\newblock {\em ISPRS Annals of the Photogrammetry, Remote Sensing and Spatial
  Information Sciences}, VIII-4/W2-2021:21--28, 2021.

\bibitem{mdpi2022}
M.W. Jahn and P.E. Bradley.
\newblock A robustness study for the extraction of watertight volumetric models
  from boundary representation data.
\newblock {\em ISPRS International Journal of Geo-Information}, 11(4):224,
  2022.

\bibitem{konde2017web}
Amol Konde and Sameer Saran.
\newblock Web enabled spatio-temporal semantic analysis of traffic noise using
  citygml.
\newblock {\em J Geomatics}, 11(2):248--59, 2017.

\bibitem{krahenmann2018high}
S~Kr{\"a}henmann, A~Walter, S~Brienen, F~Imbery, and A~Matzarakis.
\newblock High-resolution grids of hourly meteorological variables for germany.
\newblock {\em Theoretical and Applied Climatology}, 131:899--926, 2018.

\bibitem{krizhevsky2012imagenet}
Alex Krizhevsky, Ilya Sutskever, and Geoffrey~E Hinton.
\newblock Imagenet classification with deep convolutional neural networks.
\newblock {\em Advances in neural information processing systems}, 25, 2012.

\bibitem{lan2017urban}
Yuliang Lan and Qingming Zhan.
\newblock How do urban buildings impact summer air temperature? the effects of
  building configurations in space and time.
\newblock {\em Building and Environment}, 125:88--98, 2017.

\bibitem{lau2024investigating}
Tsz-Kin Lau and Tzu-Ping Lin.
\newblock Investigating the relationship between air temperature and the
  intensity of urban development using on-site measurement, satellite imagery
  and machine learning.
\newblock {\em Sustainable Cities and Society}, 100:104982, 2024.

\bibitem{ledoux2019cityjson}
Hugo Ledoux, Ken Arroyo~Ohori, Kavisha Kumar, Bal{\'a}zs Dukai, Anna Labetski,
  and Stelios Vitalis.
\newblock Cityjson: A compact and easy-to-use encoding of the citygml data
  model.
\newblock {\em Open Geospatial Data, Software and Standards}, 4(1):1--12, 2019.

\bibitem{li2021quantifying}
Huifang Li, Yanan Li, Tao Wang, Zhihua Wang, Meiling Gao, and Huanfeng Shen.
\newblock Quantifying 3d building form effects on urban land surface
  temperature and modeling seasonal correlation patterns.
\newblock {\em Building and Environment}, 204:108132, 2021.

\bibitem{lin2024does}
Anqi Lin, Hao Wu, Wenting Luo, Kaixuan Fan, and He~Liu.
\newblock How does urban heat island differ across urban functional zones?
  insights from 2d/3d urban morphology using geospatial big data.
\newblock {\em Urban Climate}, 53:101787, 2024.

\bibitem{liu2023urban}
Biao Liu, Xian Guo, and Jie Jiang.
\newblock How urban morphology relates to the urban heat island effect: A
  multi-indicator study.
\newblock {\em Sustainability}, 15(14):10787, 2023.

\bibitem{liu2020climate}
Lin Liu, Jing Liu, Lei Jin, Liru Liu, Yunfei Gao, and Xinpei Pan.
\newblock Climate-conscious spatial morphology optimization strategy using a
  method combining local climate zone parameterization concept and urban canopy
  layer model.
\newblock {\em Building and Environment}, 185:107301, 2020.

\bibitem{liu2017machine}
Lun Liu, Elisabete~A Silva, Chunyang Wu, and Hui Wang.
\newblock A machine learning-based method for the large-scale evaluation of the
  qualities of the urban environment.
\newblock {\em Computers, environment and urban systems}, 65:113--125, 2017.

\bibitem{mulder2015automatic}
DT~Mulder.
\newblock {\em Automatic Repair of 3D City Building Model Using a Voxel--based
  Repair Method}.
\newblock PhD thesis, Master Thesis, Delft Univ. Technology, Netherlands, 2015.

\bibitem{ng2012towards}
Edward Ng.
\newblock Towards planning and practical understanding of the need for
  meteorological and climatic information in the design of high-density cities:
  A case-based study of hong kong.
\newblock {\em International Journal of Climatology}, 32(4):582--598, 2012.

\bibitem{nourian2016voxelization}
Pirouz Nourian, Romulo Gon{\c{c}}alves, Sisi Zlatanova, Ken~Arroyo Ohori, and
  Anh~Vu Vo.
\newblock Voxelization algorithms for geospatial applications: Computational
  methods for voxelating spatial datasets of 3d city models containing 3d
  surface, curve and point data models.
\newblock {\em MethodsX}, 3:69--86, 2016.

\bibitem{oke1995heat}
Timothy~R Oke.
\newblock The heat island of the urban boundary layer: characteristics, causes
  and effects.
\newblock {\em Wind climate in cities}, pages 81--107, 1995.

\bibitem{padsala2021application}
Rushikesh Padsala, Ernst Gebetsroither-Geringer, Keyu Bao, and Volker Coors.
\newblock The application of citygml food water energy ade to estimate the
  biomass potential for a land use scenario.
\newblock In {\em CITIES 20.50--Creating Habitats for the 3rd Millennium:
  Smart--Sustainable--Climate Neutral. Proceedings of REAL CORP 2021, 26th
  International Conference on Urban Development, Regional Planning and
  Information Society}, pages 851--861. CORP--Competence Center of Urban and
  Regional Planning, 2021.

\bibitem{pusacker2024concept}
Klaus Pusacker, Volker Coors, J{\"o}rg-Detlef Eckhardt, and Isabel Rupf.
\newblock A concept for 3d geological and urban subsurface modeling with a
  unified voxel model examined by a case study for the city center of stuttgart
  (baden-w{\"u}rttemberg), germany.
\newblock {\em ISPRS Annals of the Photogrammetry, Remote Sensing and Spatial
  Information Sciences}, 10:193--200, 2024.

\bibitem{QGIS_software}
{QGIS Development Team}.
\newblock Qgis geographic information system.

\bibitem{raaymakers2024understanding}
Tom Raaymakers.
\newblock Understanding urban temperature differences through 2d/3d urban
  morphology.
\newblock 2024.

\bibitem{ren2020developing}
Chao Ren, Meng Cai, Xinwei Li, Yuan Shi, and Linda See.
\newblock Developing a rapid method for 3-dimensional urban morphology
  extraction using open-source data.
\newblock {\em Sustainable Cities and Society}, 53:101962, 2020.

\bibitem{ren2019capturing}
Zheng Ren, Bin Jiang, and Stefan Seipel.
\newblock Capturing and characterizing human activities using building
  locations in america.
\newblock {\em ISPRS International Journal of Geo-Information}, 8(5):200, 2019.

\bibitem{ridzuan20233d}
Nurfairunnajiha Ridzuan, Uznir Ujang, and Suhaibah Azri.
\newblock 3d vectorization and rasterization of citygml standard in wind
  simulation.
\newblock {\em Earth Science Informatics}, 16(3):2635--2647, 2023.

\bibitem{stewart2012local}
Ian~D Stewart and Tim~R Oke.
\newblock Local climate zones for urban temperature studies.
\newblock {\em Bulletin of the American Meteorological Society},
  93(12):1879--1900, 2012.

\bibitem{tanoori2024machine}
Ghazaleh Tanoori, Ali Soltani, and Atoosa Modiri.
\newblock Machine learning for urban heat island (uhi) analysis: Predicting
  land surface temperature (lst) in urban environments.
\newblock {\em Urban Climate}, 55:101962, 2024.

\bibitem{tehrani2024predicting}
Alireza~Attarhay Tehrani, Omid Veisi, Yasin Delavar, Sasan Bahrami, Saeideh
  Sobhaninia, Asma Mehan, et~al.
\newblock Predicting urban heat island in european cities: A comparative study
  of gru, dnn, and ann models using urban morphological variables.
\newblock {\em Urban Climate}, 56:102061, 2024.

\bibitem{tekouabou2022reviewing}
Stephane Cedric~Koumetio Tekouabou, El~Bachir Diop, Rida Azmi, Remi Jaligot,
  and Jerome Chenal.
\newblock Reviewing the application of machine learning methods to model urban
  form indicators in planning decision support systems: Potential, issues and
  challenges.
\newblock {\em Journal of King Saud University-Computer and Information
  Sciences}, 34(8):5943--5967, 2022.

\bibitem{venter2020hyperlocal}
Zander~S Venter, Oscar Brousse, Igor Esau, and Fred Meier.
\newblock Hyperlocal mapping of urban air temperature using remote sensing and
  crowdsourced weather data.
\newblock {\em Remote Sensing of Environment}, 242:111791, 2020.

\bibitem{vitalis2020cityjson}
Stelios Vitalis, Ken Arroyo~Ohori, and Jantien Stoter.
\newblock Cityjson in qgis: Development of an open-source plugin.
\newblock {\em Transactions in GIS}, 24(5):1147--1164, 2020.

\bibitem{voogt2003thermal}
James~A Voogt and Tim~R Oke.
\newblock Thermal remote sensing of urban climates.
\newblock {\em Remote sensing of environment}, 86(3):370--384, 2003.

\bibitem{voogt1998effects}
James~A Voogt and TR~Oke.
\newblock Effects of urban surface geometry on remotely-sensed surface
  temperature.
\newblock {\em International Journal of Remote Sensing}, 19(5):895--920, 1998.

\bibitem{wang2004image}
Zhou Wang, Alan~C Bovik, Hamid~R Sheikh, and Eero~P Simoncelli.
\newblock Image quality assessment: from error visibility to structural
  similarity.
\newblock {\em IEEE transactions on image processing}, 13(4):600--612, 2004.

\bibitem{willenborg2016semantic}
BRUNO Willenborg, Maximilian Sindram, and THOMAS~H Kolbe.
\newblock Semantic 3d city models serving as information hub for 3d field based
  simulations.
\newblock {\em L{\"o}sungen f{\"u}r eine Welt im Wandel}, pages 54--65, 2016.

\bibitem{wu2013development}
Chih-Da Wu, Shih-Chun~Candice Lung, and Jihn-Fa Jan.
\newblock Development of a 3-d urbanization index using digital terrain models
  for surface urban heat island effects.
\newblock {\em ISPRS Journal of Photogrammetry and Remote Sensing}, 81:1--11,
  2013.

\bibitem{wu2010urban}
Jianguo Wu.
\newblock Urban sustainability: an inevitable goal of landscape research, 2010.

\bibitem{yang2020application}
Zhiwei Yang, Yingbiao Chen, Zihao Zheng, Qingyao Huang, and Zhifeng Wu.
\newblock Application of building geometry indexes to assess the correlation
  between buildings and air temperature.
\newblock {\em Building and Environment}, 167:106477, 2020.

\bibitem{yoo2018challenges}
Sung~J Yoo, Taeyong Kwon, and Young~S Lyoo.
\newblock Challenges of influenza a viruses in humans and animals and current
  animal vaccines as an effective control measure.
\newblock {\em Clinical and experimental vaccine research}, 7(1):1--15, 2018.

\bibitem{zhang2018unreasonable}
Richard Zhang, Phillip Isola, Alexei~A Efros, Eli Shechtman, and Oliver Wang.
\newblock The unreasonable effectiveness of deep features as a perceptual
  metric.
\newblock In {\em Proceedings of the IEEE conference on computer vision and
  pattern recognition}, pages 586--595, 2018.

\bibitem{zheng2019higher}
Zhong Zheng, Weiqi Zhou, Jingli Yan, Yuguo Qian, Jia Wang, and Weifeng Li.
\newblock The higher, the cooler? effects of building height on land surface
  temperatures in residential areas of beijing.
\newblock {\em Physics and Chemistry of the Earth, Parts A/B/C}, 110:149--156,
  2019.

\bibitem{zhong2017revealing}
Chen Zhong, Markus Schl{\"a}pfer, Stefan M{\"u}ller~Arisona, Michael Batty,
  Carlo Ratti, and Gerhard Schmitt.
\newblock Revealing centrality in the spatial structure of cities from human
  activity patterns.
\newblock {\em Urban Studies}, 54(2):437--455, 2017.

\bibitem{zhou2011does}
Weiqi Zhou, Ganlin Huang, and Mary~L Cadenasso.
\newblock Does spatial configuration matter? understanding the effects of land
  cover pattern on land surface temperature in urban landscapes.
\newblock {\em Landscape and urban planning}, 102(1):54--63, 2011.

\bibitem{zlatanova2016towards}
Sisi Zlatanova, Pirouz Nourian, Romulo Gon{\c{c}}alves, and Anh~Vu Vo.
\newblock Towards 3d raster gis: On developing a raster engine for spatial
  dbms.
\newblock In {\em ISPRS WG IV/2 Workshop: Global Geospatial Information and
  High Resolution Global Land Cover/Land Use Mapping}, 2016.

\end{thebibliography}

\end{document}